# Experiments on Turkish ASR with Self-Supervised Speech Representation Learning

Ali Safaya, Engin Erzin
KUIS AI Center
Computer Engineering Department, Koç University

## Abstract

While the Turkish language is listed among low-resource languages, literature on Turkish automatic speech recognition (ASR) is relatively old. In this report, we present our findings on Turkish ASR with speech representation learning using HUBERT. We investigate pre-training HUBERT for Turkish with large-scale data curated from online resources. We pre-train our model using 6,500 hours of speech data from YouTube. The results show that the models are not ready for commercial use since they are not robust against disturbances that typically occur in real-world settings such as variations in accents, slang, background noise and interference. We analyze typical errors and the limitations of the models for use in commercial settings.

## Introduction

As the demand for expanding speech processing models to a broader mass of languages grows, labeled-data scarcity is becoming a fundamental problem for low-resource languages. Unsupervised learning approaches, which require much less labeled data, can help to solve this problem. This effect can be seen in Natural Language Processing (NLP) using unsupervised language modeling (ELMO, BERT) [1, 2]. Likewise, speech representation learning models have been developed for speech tasks. For instance, WAV2VEC and WAV2VEC 2.0, are early approaches to speech representation learning using contrastive learning [3, 4]. Methods such as Discrete-BERT, Hidden-unit BERT (HUBERT), and WAVLM utilize BERT-style Masked Hidden Unit Prediction as an objective for self-supervised training [5-7]. These models showed significant improvements over supervised methods on speech tasks such as Automatic Speech Recognition (ASR) and other tasks by utilizing unlabeled large-scale data [3-5].

Unfortunately, large-scale data such as LibriSpeech [8], Librilight [9] are mainly available for the English language. Due to this, most speech representation models were initially pre-trained for English, which puts English at a big advantage compared to other languages. More recently, speech models were extended to multilingual settings, as in CPC-8K and XLS-R [10-11]. However, these models do not cover all languages equally in their pre-training data. For instance, the Turkish subset of XLS-R pre-training data contained only 70 hrs of audio.

In this work, we present experiment results for monolingual speech transcription in Turkish based on the well known HUBERT architecture [6]. We pre-train HUBERT using 6.5k hours of unlabeled data from YouTube and then fine-tune it on various datasets. Our analyses show the models lack the robustness required for commercial ASR applications.

## Turkish ASR

Turkish ASR still has room for improvement and has yet to catch up to English ASR performance. Previous work on Turkish ASR has been limited due to a lack of resource variability [12]. Early work on Turkish ASR utilizes Hidden Markov Models (HMMs) and Gaussian Mixture Models (GMMs) [12]. For instance, a GMM-based system achieved a word error rate of 27.70% on the Turkish Broadcast News (TBN) dataset [13].

With the rise of neural networks, acoustic neural methods such as DeepSpeech2 [14] introduced notable improvements to Turkish ASR [13, 15]. Another recent study examined neural network-based acoustic and language models for the Turkish ASR [16]. They utilized time-delay neural networks (TDNNs) for the acoustic model, using both cross-entropy and sequence discriminative objective functions. By incorporating an LSTM-based language model, they attained a 9.83% word error rate on the TBN dataset, which is approximately halved compared to earlier HMM/GMM-based models. Attempts to deal with limited data problems include utilizing language models in variant ways [17, 18], and applying data augmentation [15] to improve ASR performance. In this work, we approach Turkish ASR by taking advantage of self-supervised learning with pre-training on 6.5K hours of unlabeled audio from YouTube.

## Development

HUBERT is a self-supervised speech representation model that is pre-trained on predicting pseudo-labels of masked audio segments. These pseudo-labels are automatically generated using audio features, such as MFCCs or by clustering hidden representations [6]. In this section, we explain the process of pre-training HUBERT for Turkish.

### Model

Following the original HUBERT paper [6], we pre-train a set of three models BASE / LARGE / XLARGE. HUBERT encodes speech using two components: the feature extractor and the feature encoder. The feature extractor is composed of a set of consecutive Convolutional blocks. This feature extractor takes in raw speech signal with a sample-rate of 16,000 Hz, and encodes it as a sequence of hidden vectors. Each hidden vector encodes 20ms of audio. Afterward, the extracted hidden vectors are passed into the feature encoder, which is a Transformer encoder [19]. We employ seven consecutive Convolutional blocks in the feature extractor. Convolutional blocks consist of a one-dimensional convolution layer of 512 channels, with five strides in the first block and two in the rest. A Layer Normalization [20] and GELU [21] activation follow the convolution layer. This configuration is the same for all model sizes. We configure the feature encoder's architecture depending on the model's size.

### Data

English speech models and multi-lingual models were pre-trained using relatively big speech datasets, such as LibriSpeech (960 hours) [8], Librilight (60K hours) [9] and mixtures that go up to 436K hours of speech. However, these do not cover all languages equally, and the Turkish portion of such sets are only 70 hours. To tackle this, we build a new pre-training set from YouTube videos to solve the scarcity of Turkish speech data. Overall, we download 6,500 hours of videos and extract the audio content.

### Pre-training

HUBERT is pre-trained using a self-supervised method inspired by the masked language modeling approach used in BERT. Instead of predicting the masked token as in BERT, HUBERT is pre-trained to predict the hidden-units of the masked audio frames. We start pre-training by randomly picking a portion of the input frames, and we mask corresponding spans. Next, we pre-train HUBERT on predicting the hidden-units of audio frames.

## Performance Analysis

To assess the quality of the learned speech representations, we fine-tune our models on various Turkish ASR datasets using Connectionist Temporal Classification (CTC) loss [22].

We conduct a qualitative analysis on the predictions produced by our models, and include example errors from models' outputs on a test sample taken from COMMONVOICE dataset in Table 1. We find that the models are prone to produce trivial spelling errors.

Moreover, we observe higher error-rates in shorter speech segments, specifically with segments with three words and less. This is especially an important observation for keyword-spotting and intent recognition, which are essential tasks for ASR applications in banking, customer call centers and telecommunication fields. We investigate this empirically and we witness 10% and 6% relative increase in Word Error Rates, for BASE and LARGE models relatively, on short segments from CommonVoice dataset.

Another crucial aspect is the bias of the models towards news as a domain. Since the majority of the pre-training and fine-tuning data comes from YouTube videos, this reduced the models' ability to generalize to other domains. Furthermore, we find that the models are heavily tuned towards politics related words: `almanya, durum, kurul, birleşmiş, örgüt, politika, yetkililer, destek`.

Due to the formality and clarity of pre-training and fine-tuning speech data sets, the resulting models have a high tendency towards generating erroneous outputs when they are tested with inputs which contain background noise, dialected Turkish speech, or heavy regional accents.

Table 1 - Test samples with the transcriptions predicted by our models

| Reference | muhalefet bu sava inanmıyor |
|---|---|
| BASE | mahale fet usava inanmi or |
| LARGE | muhalefet pusava inanmı yoğ |
| Reference | damarlarında dolaşan kan değil yanardağ lavlarıydı sanki |
| BASE | damarunında oluaşam kan değil yanar dağılan larıydi |
| LARGE | damarlarında dolaşan kan değil yanarda ağılanlarıydı |